\pdfoutput=1

\documentclass[11pt]{article}

\usepackage[]{acl}
\usepackage{adjustbox}
\usepackage[acronym]{glossaries}

\usepackage{times}
\usepackage{latexsym}

\usepackage[T1]{fontenc}

\usepackage[utf8]{inputenc}

\usepackage{microtype}

\title{ML\_LTU at SemEval-2022 Task 4: T5 Towards Identifying Patronizing and Condescending Language}

\author{Tosin Adewumi, \ Lama Alkhaled, \ Hamam Mokayed, \ Foteini Liwicki \and Marcus Liwicki \\
         Machine Learning Group \\ EISLAB, SRT \\ Luleå University of Technology\\
         firstname.lastname@ltu.se}

\makeglossaries

\newacronym{nlp}{NLP}{Natural Language Processing}
\newacronym{ner}{NER}{Named Entity Recognition}
\newacronym{sa}{SA}{Sentiment Analysis}
\newacronym{bow}{BoW}{bag-of-words}
\newacronym{cbow}{CBoW}{continuous Bag-of-Words}
\newacronym{sltc}{SLTC}{Swedish Language Technology Conference}
\newacronym{ann}{ANN}{artificial neural network}
\newacronym{nn}{NN}{neural network}
\newacronym{lstm}{LSTM}{Long Short Term Memory Network}
\newacronym{sota}{SoTA}{state-of-the-art}
\newacronym{nlg}{NLG}{Natural Language Generation}
\newacronym{mwe}{MWE}{Multi-Word Expression}
\newacronym{cnn}{CNN}{Convolutional Neural Network}
\newacronym{pcl}{PCL}{patronizing and condescending language}
\newacronym{mt}{MT}{Machine Translation}
\newacronym{gdc}{GDC}{Gothenburg Dialog Corpus}
\newacronym{t5}{T5}{Text-to-Text-Transfer Transformer}
\newacronym{roberta}{RoBERTa}{Robustly optimized BERT approach}
\newacronym{bert}{BERT}{Bidirectional Encoder Representations from Transformers}
\newacronym{mcc}{MCC}{Matthews Correlation Coefficient}
\newacronym{ai}{AI}{Artificial Intelligence}
\newacronym{xai}{XAI}{explainable artificial intelligence}
\newacronym{lime}{LIME}{Local Interpretable Model-agnostic Explanations}
\newacronym{bilstm}{Bi-LSTM}{Bi-Directional Long Short Term Memory Network}
\newacronym{rnn}{RNN}{Recurrent Neural Network}
\newacronym{ltu}{LTU}{Luleå University of Technology}

\begin{document}
\maketitle
\begin{abstract}
This paper describes the system used by the Machine Learning Group of \acrshort{ltu} in subtask 1 of the SemEval-2022 Task 4: Patronizing and Condescending Language (PCL) Detection.
Our system consists of finetuning a pretrained \acrfull{t5} and innovatively reducing its out-of-class predictions.
The main contributions of this paper are 1) the description of the implementation details of the \acrshort{t5} model we used, 2) analysis of the successes \& struggles of the model in this task, and 3) ablation studies beyond the official submission to ascertain the relative importance of data split.
Our model achieves an F1 score of 0.5452 on the official test set.
\end{abstract}

\citet{perezalmendros2020dont} introduced the dataset for the SemEval-2022 Task 4 \cite{perezalmendros2022semeval}\footnote{semeval.github.io/SemEval2022/tasks}.
The dataset covers the English language.
It is meant to support \acrfull{nlp} models in identifying \acrshort{pcl} towards vulnerable communities, such as poor families and refugees.
The dataset is designed for 2 subtasks in the competition.
Subtask 1 is a binary classification task of predicting the presence of \acrshort{pcl} while subtask 2 is a multi-label classification task of predicting \acrshort{pcl} categories.
We address subtask 1 in this system paper.

\acrshort{pcl} is an expression that depicts someone in a compassionate way or shows a superior attitude of the speaker \cite{perezalmendros2022semeval}.
\acrshort{pcl} identification is important because \acrshort{pcl} has been shown to have harmful effects on vulnerable groups \cite{fox1996interability,morris2007patronizing,bell2013raising,wang-potts-2019-talkdown}.
This task of identifying and categorizing \acrshort{pcl} is apparently more challenging than some other types of harmful language because it is subtle and generally used with good intentions \cite{wang-potts-2019-talkdown, gilda2022predicting}.

The main strategy of our system, to address the challenge, was to use a recent \acrshort{sota} model (\acrshort{t5}) in a simple, novel way to reduce out-of-class predictions.
We discovered that our system achieves a relatively good performance on the task and \acrshort{pcl} identification is a challenging task, due to its subtle nature.
It achieved an F1 score of 0.5452 on the test set while the best score was 0.651.
This made us rank 27 (66th percentile) out of 78 and we surpass the official \acrshort{roberta} baseline.
We perform error analysis and ablation studies to evaluate the strengths and weaknesses of the model.
We contribute the model checkpoint publicly on the HuggingFace hub\footnote{huggingface.co/tosin/pcl\_22} and the \acrshort{t5} code \footnote{github.com/tosingithub/pcl (after another competition)}

The rest of this paper is organized as follows.
Section \ref{background} gives a brief background of related work in \acrshort{pcl}.
Section \ref{overview} gives the system overview of what we used for the task.
Section \ref{exp} describes the experimental setup for the task and the additional experiments beyond the official submission.
Section \ref{results} gives the tables of results and discusses relevant observations from the results.
We share concluding remarks in section \ref{conclusion}.

\section{Background}
\label{background}
Work on various sorts of harmful language in \acrshort{nlp} has mostly concentrated on explicit aggressive and brazen phenomena \cite{perezalmendros2022semeval}.
Scholars are striving to distinguish between harmful and unhealthy language by identifying the fundamental characteristics of unhealthy language.
\citet{price-etal-2020-six} proposed one of the most recent efforts in this regard.
The research introduced a new dataset containing 44,000 comments with the unhealthy category sub-classified as either (1) hostile; (2) antagonistic, insulting, provocative or trolling; (3) dismissive; (4) condescending or patronizing; (5) sarcastic; and/or (6) an unfair generalisation.
In their work, it is assumed that the language with a \acrshort{pcl} tone will assume an attitude of superiority, implying that the other speakers/listeners are ignorant, naive, or unintelligent.
In such scenarios, the language will usually imply that the other speaker should not be taken seriously.

Similarly, \citet{morris2007patronizing} explains the high likelihood of using \acrshort{pcl} language when there is discussion between two persons with different mental health conditions.
He demonstrated that patronizing language is common when a discussion occurred between a cashier with no cognitive issue and a customer who suffers from cognitive disability.
Overall, \acrshort{pcl} does not have an obvious negative or critical language and there is the challenge of limited, high-quality labelled data.  

There have been different efforts at automatically detecting \acrshort{pcl}.
\citet{wang-potts-2019-talkdown} showed that models with contextual representations are much better at identifying \acrshort{pcl} and this bolstered the hypothesis that context is essential for \acrshort{pcl} detection.
They implemented the \acrshort{bert} model, which deploys a Transformer-based encoder architecture, on the TALKDOWN corpus they introduced.
Both the base and large versions of the \acrshort{bert} model are implemented and evaluated over the new proposed corpus for balanced and imbalanced data.
\citet{price-etal-2020-six} added more context to their work by comparing the performance of \acrshort{bert} to human performance in order to better understand the model's performance.
In their experiments, they observed that the \acrshort{bert} model detects \acrshort{pcl} with a 78\% accuracy, whereas the average over human annotators does so with a 72\% accuracy.
\cite{warholm2021detecting} also finetuned a \acrshort{bert} model to classify the unhealthy comments in Norwegian data.
This model was subjected to a variety of finetuning approaches to distinguish between condescending and non-condescending cases and in the binary classification subtask, the best accuracy was 0.862.

\subsection{Data}
“Don’t patronize me” is an annotated dataset of \acrshort{pcl} by \cite{perezalmendros2020dont} through crowdsourcing.
It is a collection of texts which targets vulnerable communities.
The dataset is extracted from News On Web (NoW) corpus\footnote{english-corpora.org/now/}, containing web articles from over 20 English-speaking countries.
It contains 10,637 paragraphs.
In addition to the words (\textit{disabled, homeless, hopeless, immigrant, in need, migrant, poor families, refugee, vulnerable and women}) for identifying \acrshort{pcl} for annotation in paragraphs, the following traits are also identified as indicators and used for acquiring the dataset:
\begin{itemize}
    \item Words expressing feeling of pity towards the vulnerable community. For example: \textit{god bless the victims , all those people and their poor families , and i feel so sorry but i want to tell them it was n't my son who did this , it was a different seifeddine}
    \item Words describing the vulnerable community as lacking certain privileges, knowledge or experience. For example: \textit{After Vatican controversy, McDonald’s helps feed homeless in Rome}
    \item Expressions that present members of the vulnerable community as victims. For example: \textit{the biggest challenge is the no work policy . i think that refugees who come here , or asylum seekers , they 're unable to work and they have kids here -- their kids are stateless . that 's really the cause of a lot of stress in the community}
\end{itemize}

The dataset was annotated by 3 expert annotators.
It has two-level classification of PCL: binary classification used to determine if a paragraph has \acrshort{pcl} or not, and then categorical label for those with \acrshort{pcl}.
The categorical classification has three higher-level categories: saviour, expert and poet.
"Other" category is the final category to classify all paragraphs with \acrshort{pcl} but that do not fit any of the previous categories. 
The saviour category represents text in which the author is in a privileged class as opposed to the target community.
It has two subcategories: unbalanced power relations and Shallow solutions.
The expert category is for text where the author is also in a privileged position and presents themselves as knowing better than the target group what their needs are.
It also has two subcategories: presupposition and authority voice.
The final category “Poet” is identified by how the author frames the community with a literary style writing.
It has three subcategories: Metaphor, Compassion and The poorer the merrier.

\section{System Overview}
\label{overview}

The \acrshort{t5} architecture \cite{JMLR:v21:20-074} is very similar to the originally proposed architecture of the Transformer by \citet{vaswani2017attention}.
We use the pretrained base version of the  model from the HuggingFace hub \cite{wolf-etal-2020-transformers}.
Input sequence of tokens are mapped to embeddings and then passed to the encoder, which has alternating set of multi-head attention and feed-forward layers.
The attention mechanism \cite{bahdanau2015neural} replaces each element of a sequence by a weighted average of the remaining sequence \cite{JMLR:v21:20-074}.
In addition to each self-attention layer of the decoder, there is the standard attention mechanism.
As self-attention is order-independent, relative position embeddings are used in the architecture.

The training method (for both pretraining and finetuning) uses maximum likelihood objective (i.e. teacher forcing) and a cross entropy loss \cite{JMLR:v21:20-074}.
The model was pretrained on 34B tokens.
Adam optimizer is used for optmization during finetuning.
The model has 12 layers each in the encoder and decoder blocks and a total of 220M parameters \cite{JMLR:v21:20-074}.
When we refer to \acrshort{t5}, we mean the base model, except where explicitly stated otherwise.
The size of the model meant that a batch size of 64 or 32 required more memory than what is available on a single V100 GPU, so we lowered the batch size to 16.
\acrshort{t5} takes a hyperparameter called a task prefix.
We, hence, use `classification: ' as the task prefix.

\begin{figure}[h]
\centering
\includegraphics[width=0.4\textwidth]{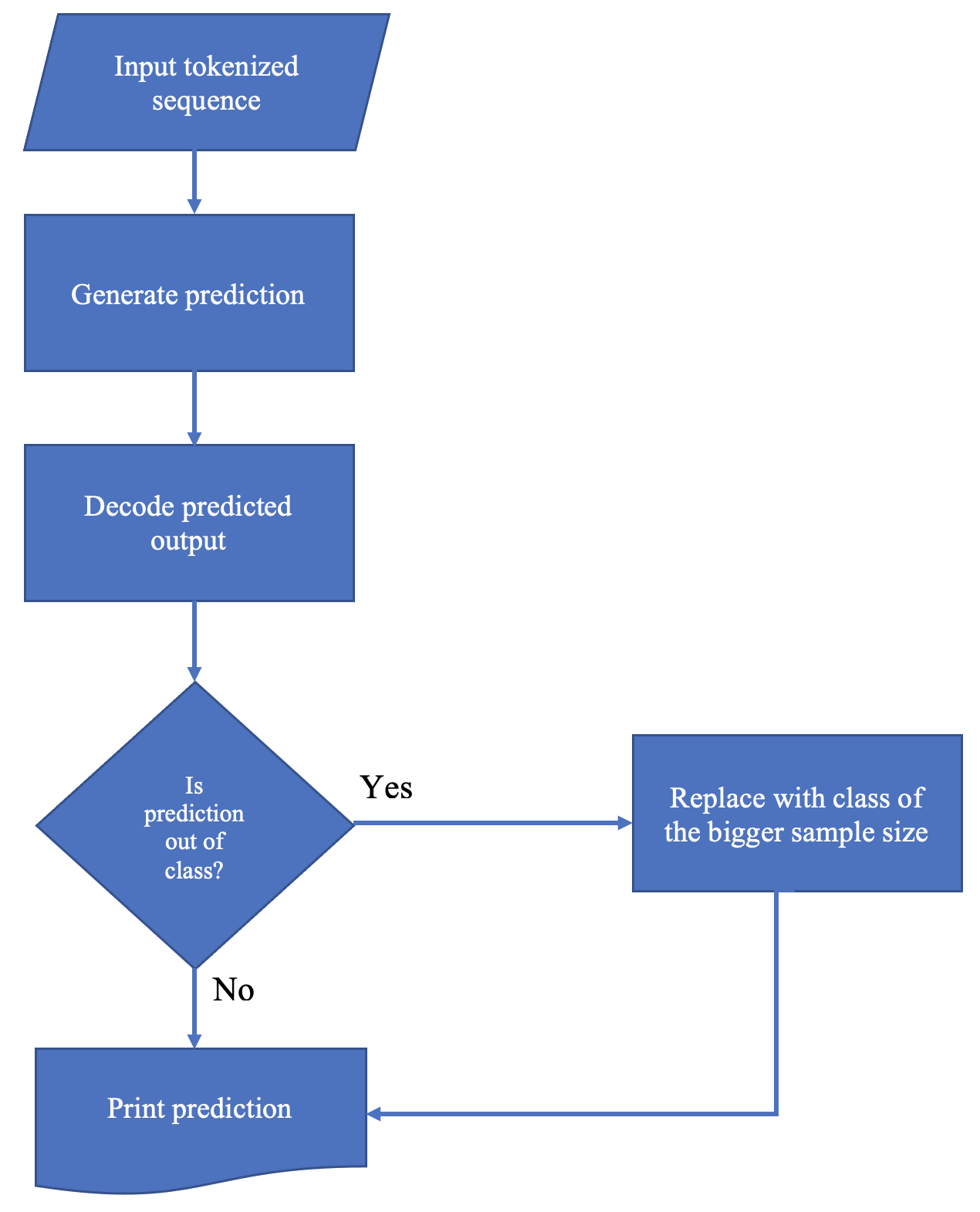}
\caption{Flowchart of out-of-class code section for the  \acrshort{t5} model during prediction.}
\label{fig:flow}
\end{figure}

We introduced a correction to the out-of-class prediction of the model, as shown in the flow chart in Figure \ref{fig:flow}.
\citet{JMLR:v21:20-074} mentioned this issue as a possibility but they did not experience it.
The issue appears to be because all the tasks the \acrshort{t5} model is trained on are framed as "text-to-text" before training.
Hence, sometimes, the model might predict tokens seen during training but that do not belong to the category of classes in a classification task.
This behaviour seems more common in the initial epochs of training and may not even occur sometimes.
We further observed that replacing target labels with numbers and explicitly typecasting them as string reduces this occurrence, as the model becomes more stable with predictions.

We split 10\% of the training set for validation (dev set) for both of our submissions to the competition.
We explored different sizes, however, in further ablation studies, as explained in the next section.
The 2 submissions of prediction files are based on 2 adaptive optimizers: Adam and AdamW \cite{loshchilov2019decoupled}.
The predictions based on Adam had the better F1 score.
Each experimental run was for 3 epochs and the model checkpoint with the lowest validation loss was saved and used to make prediction on the test set.
The initial learning rate and scheduler for both submissions are 2e-4 and linear schedule with warmup, respectively.

\section{Experimental Setup}
\label{exp}

All the experiments were conducted on a shared DGX-1 cluster of 8 × 32GB Nvidia V100 GPUs.
The server runs on Ubuntu 18 OS and has 80 CPU cores.
The experiments were conducted in a Python (3.6.9) virtual environment with the PyTorch framework (1.8.1+cu102).
We use both the training \& test data provided by \citet{perezalmendros2020dont}.
Besides the 2 submissions of prediction files, we perform ablation studies over the training/dev set split ratio (95\%/5\%, 90\%/10\%, 85\%/15\%, and 80\%/20\%).
The training set was shuffled before splitting each dev set.
We evaluate all the models using macro F1 scores, precision (P) and recall (R).
In the absence of the ground truth of the test set, we perform error analysis by constructing the confusion matrix on a split of the dev set (20\%).
Further to that, in order to have a basis of comparison of the \acrshort{t5} model's strengths and struggles with the official \acrshort{roberta} baseline, we removed the 10 examples provided in Table 5 by \citet{perezalmendros2022semeval} from the training set and concatenated them with the dev set before training and evaluation.
The predictions of 9 of the samples are given in Table \ref{table:error}.

Evaluation of the available data, by code, before and after running the script provided by \citet{perezalmendros2022semeval} to categorize the labels into 0 (neg) and 1 (pos) (for subtask 1) reveals that there are a total of 10,469 samples.
The script treated paragraphs with the original labels 0 and 1 as 0  (instances not containing \acrshort{pcl}) and paragraphs with the original labels 2, 3 and 4 as 1 (instances containing \acrshort{pcl}).
After running the script, the following are obtained: 9,476 samples classified as 0 and 993 classified as 1 in the training set.
The test set has 3,832 samples.
Before training, the following preprocessing steps were applied to all splits of the data: 

\begin{itemize} 

\item Emails \& URLs are removed. 

\item All the characters are made lowercase. 

\item Extra spaces are removed.

\item Special characters such as hashtags(\#) and emojis are removed. 

\item Numbers \& IP addresses are removed. 

\end{itemize}

\section{Results and Discussion}
\label{results}

Our model performed relatively well with an F1 score of 0.5452 in the official assessment.
This made it rank 27 (the 66th percentile) out of the 78 scores.
All the F1 scores we report are macro scores.
Our model has 11\% advantage over the  \acrshort{roberta} baseline, which achieved 0.4911, as shown in Table \ref{abridged}.
Indeed, our second submission, based on the AdamW optimizer, also performs better than the baseline, achieving an F1 score of 0.5282, precision and recall of 0.5976 and 0.4732, respectively.
The \acrshort{t5} model may have performed even better in the official rankings but for the shortcoming we described in section \ref{overview}.
In ablation studies, as shown in Table \ref{ablation}, we observe that training/dev set split ratio affects the performance of the system.
All the results are based on submissions to the official evaluation system\footnote{competitions.codalab.org/competitions/34344}.
Using 5\% of the training set as the dev set gave the worst F1 score but we observe improvements as the size is increased, though not linearly.
We observe a sharp rise in F1 score when we increase the split from 5\% to 10\% but the rate of increase falls for subsequent increases.

\begin{table}[h!]
\centering
\resizebox{\columnwidth}{!}{%
\begin{tabular}{lllll}
\hline
Model & Rank & P & R & F1\\
\hline
best & 1 & 0.646 & 0.6562 & 0.651\\
\acrshort{t5} (ours) & 27 & 0.5801 & 0.5142 & 0.5452 \\
\acrshort{roberta} baseline & 43 & 0.3935 & 0.653 & 0.4911 \\
worst & 78 & 0.1059 & 0.0284 & 0.0448 \\
\hline
\end{tabular}
}
\caption{\label{abridged}
Abridged official result ranking for subtask 1.
}
\end{table}

\begin{table}[h!]
\centering
\resizebox{\columnwidth}{!}{%
\begin{tabular}{llll}
\hline
Model (dev split) & P & R & F1\\
\hline
\acrshort{t5} (5\%) & 0.0725 & 0.8643 & 0.1339 \\
\acrshort{t5} (10\%) & 0.6725 & 0.3628 & 0.4713 \\
\acrshort{t5} (15\%) & 0.6067 & 0.4574 & 0.5216 \\
\acrshort{t5} (20\%) & 0.5818 & 0.5047 & 0.5405 \\
\hline
\end{tabular}
}
\caption{\label{ablation}
Ablation studies results on the test set for subtask 1.
Hyperparameters are the same for all model modifications. The \acrshort{t5} (10\%) model is retrained afresh like the others, to avoid test/dev set feedback because of the samples in table \ref{table:error}.
}
\end{table}

\subsection{Error Analysis}
\begin{figure}[h]
\centering
\includegraphics[width=0.4\textwidth]{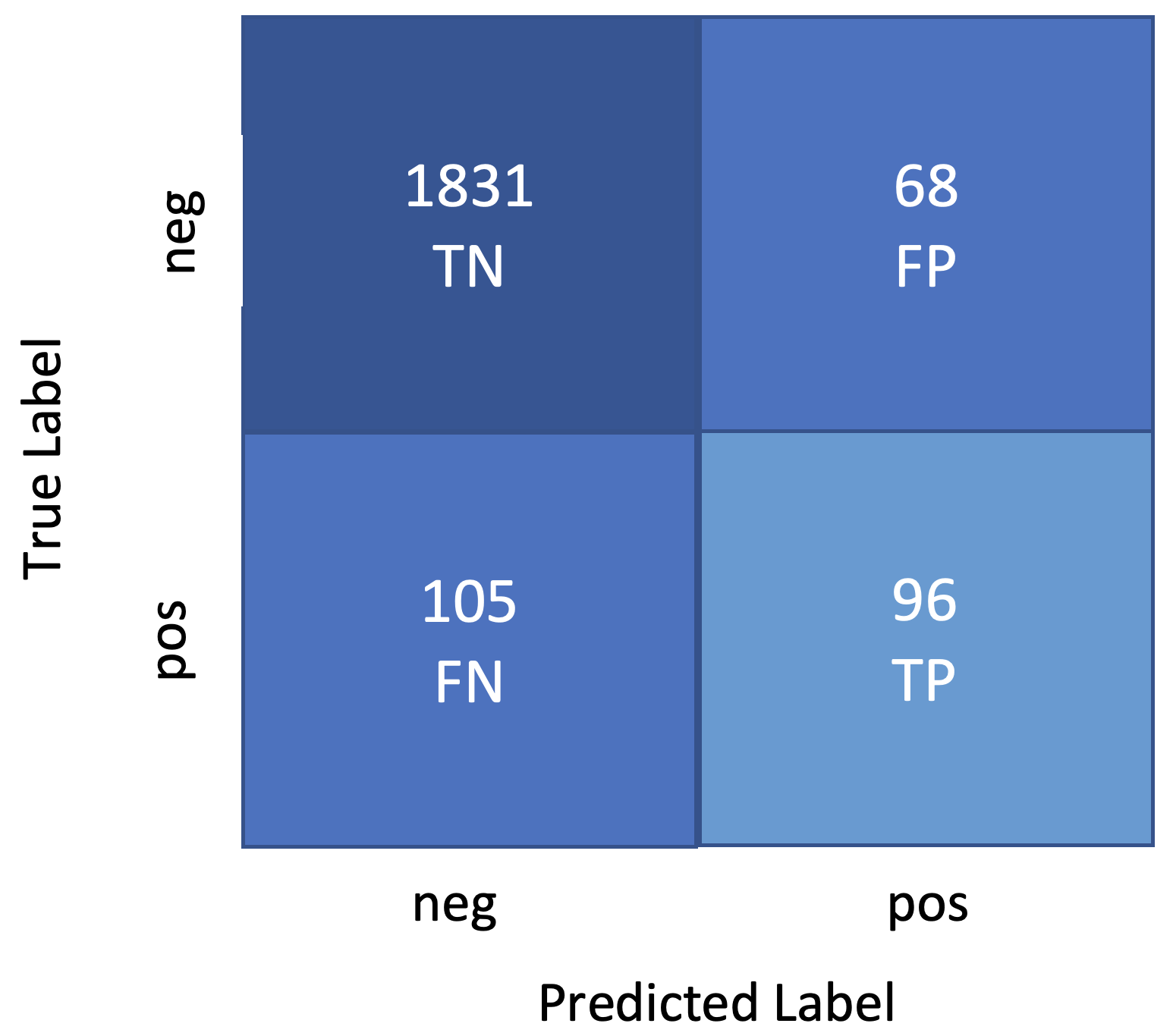}
\caption{Confusion matrix of \acrshort{t5} (20\%) on the dev set (20\%).
Macro F1 (0.7405): [0.9549 (neg) 0.5260 (pos)]}
\label{fig:cmat}
\end{figure}

Since the ground truth labels of the test set are not available, we perform error analysis on the dev set.
The \acrshort{t5} (20\%) model achieves an F1 score of 0.7405 on the dev set (20\%).
However, the confusion matrix, as depicted in Figure \ref{fig:cmat}, reveals that the model predicted 0 (neg) correctly 96.4\% of the time while struggling to make the correct predictions when it came to 1 (pos), making only 47.8\% of predictions correctly.
This is very likely due to data imbalance, as 90.5\% of the total training set contains samples labeled as 0 (neg).
Ways of mitigating this may include data augmentation, possibly in a similar strategy to that used by \citet{sabry2022hat5}, where an autoregressive model was deployed \cite{adewumi2021sm}.
A more careful stratification of the data split may also be helpful in this case.

\citet{perezalmendros2022semeval} report that the models they considered struggled to detect certain categories of \acrshort{pcl}.
We observe a similar challenge though our model achieves a better performance than the official baseline.
For example, our \acrshort{t5} (20\%) model's predictions for the same examples shown by \citet{perezalmendros2022semeval} for subtask 1 reveal that our model correctly predicts 5 out of the 9 displayed in Table \ref{table:error}, unlike the 3 correct predictions out of the 10 by the official baseline.
The reason the \acrshort{t5} (20\%) may have misclassified 2 of the samples labeled 0 (neg) in Table \ref{table:error} may be because of tokens such as \textit{vulnerable patients} and \textit{hopelessly}, since they belong to the keywords used for annotating paragraphs with \acrshort{pcl}, as discussed in section \ref{background}.

\begin{table}[h!]
\centering
\resizebox{\columnwidth}{!}{%
\begin{tabular}{p{0.10\linewidth} p{0.80\linewidth}p{0.10\linewidth}}
\hline
Pred. & Paragraph & Gold
\\
\hline
pos & from his personal story and real-life encounters with poor families , manpower correspondent toh yong chuan suggested shifting the focus from poor parents who repeatedly make bad decisions to their children ( """" lifting families out of poverty : focus on the children ; last thursday ) . & pos
\\ 
\hline
pos & he said their efforts should not stop only at creating many graduates but also extended to students from poor families so that they could break away from the cycle of poverty & pos
\\
\hline
neg & smyth told hkfp : """" the biggest challenge is the no work policy . i think that refugees who come here , or asylum seekers , they 're unable to work and they have kids here -- their kids are stateless . that 's really the cause of a lot of stress in the community . & pos
\\
\hline
neg & the people of khyber pakhtunkhwa are resilient . i did not see hopelessness on any face , """" he said . & pos
\\
\hline
pos & teach kids to give back : when kang runs summer camps with kids , she includes """" contribution fridays """" -- the kids work together as a team to make sandwiches for the homeless and dole out the food in shelters . & pos
\\ 
\hline
pos & these shocking failures will continue to happen unless the government tackles the heart of the problem -- the chronic underfunding of social care which is piling excruciating pressure on the nhs , leaving vulnerable patients without a lifeline . & neg
\\
\hline
neg & lilly-hue : his ability to make sure our family is never in need -- his sacrificial self . & neg
\\
\hline
pos & "any kenyan small-scale farmer with such an income could not be said to be hopelessly mired in agrarian destitution . but of course , nothing in life is ever so simple as to allow for neat and precise answers ." & neg
\\
\hline
neg & "selective kindness : in europe , some refugees are more equal than others" & neg
\\
\hline
\end{tabular}
}
\caption{\label{table:error}Example predictions by \acrshort{t5} based on the dev set}
\end{table}

\section{Conclusion}
\label{conclusion}

We describe the system involving the pretrained \acrshort{t5} model, which we use for our submission for the subtask 1 of the SemEval-2022 Task 4.
We split 10\% of the training set as dev set for hyperparameter evaluation in our official submission.
Typecasting integer values, which represent classes, as string before feeding the \acrshort{t5} model and adjusting for out-of-class predictions improved the stability of the model in making predictions.
Furthermore, in the post-competition phase, we performed ablation studies on the relative importance of dataset split by experimenting with different ratios of the training/dev set and showed what the model struggles with.
Our results show that the encoder-decoder \acrshort{t5} model is competitive in this binary task and can obtain better performance with more hyperparameter tuning.

\section*{Acknowledgements}
The authors wish to thank the anonymous reviewers for their feedback and the task organisers for their prompt attention whenever it was required.

\bibliography{custom}
\bibliographystyle{acl_natbib}

\printglossary[type=\acronymtype]

\end{document}